\documentclass{INTERSPEECH2023}

\interspeechcameraready

\title{GhostRNN: Reducing State Redundancy in RNN with Cheap Operations}
\name{Hang Zhou$^{1,2}$, Xiaoxu Zheng$^3$, Yunhe Wang$^{2*}$, Michael Bi Mi$^3$, Deyi Xiong$^{1,4*}$\thanks{*Corresponding author.}, Kai Han$^{2}$}
\address{
  $^1$ College of Intelligence and Computing, Tianjin University, Tianjin, China \\
  $^2$ Huawei Noah's Ark Lab, China\\
  $^3$ Huawei International Pte Ltd, Singapore\\
  $^4$ School of Computer Science and Technology, Kashi University, Kashi, China}
\email{ \{zhouhang25, zhengxiaoxu1, yunhe.wang, Michael.Bi.Mi\}@huawei.com,  dyxiong@tju.edu.cn, kai.han@huawei.com}
\usepackage{color}
\usepackage{float}
\usepackage{changes}
\usepackage{color}
\usepackage{caption}
\usepackage{subcaption}
\usepackage{ulem}
\usepackage{bm}
\usepackage{amsmath}
\usepackage{hyperref}
\usepackage{threeparttable}
\begin{document}

\maketitle
 
\begin{abstract}
	Recurrent neural network (RNNs) that are capable of modeling long-distance dependencies are widely used in various speech tasks, eg., keyword spotting (KWS) and speech enhancement (SE). Due to the limitation of power and memory in low-resource devices, efficient RNN models are urgently required for real-world applications. In this paper, we propose an efficient RNN architecture, GhostRNN, which reduces hidden state redundancy with cheap operations. In particular, we observe that partial dimensions of hidden states are similar to the others in trained RNN models, suggesting that redundancy exists in specific RNNs. To reduce the redundancy and hence computational cost, we propose to first generate a few \emph{intrinsic} states, and then apply cheap operations to produce \emph{ghost} states based on the \emph{intrinsic} states. Experiments on KWS and SE tasks demonstrate that the proposed GhostRNN significantly reduces the memory usage ($\sim$40\%) and computation cost while keeping performance similar. 	

\end{abstract}

\section{Introduction}\label{intro}

Recent years have witnessed that substantial improvements have been made in a wide range of speech tasks with the rapid development of neural networks. Among these neural networks, RNNs, e.g., LSTMs \cite{hochreiter1997long} or GRUs \cite{chung2014empirical}, are widely employed in various speech-related tasks in low-resource devices (e.g., mobile phones), such as KWS \cite{zhang2017hello, rybakov2020streaming}, SE \cite{hu2020dccrn, luo2020dual}, automatic speech recognition \cite{rao2017exploring}, acoustic echo cancellation \cite{pfeifenberger2020nonlinear, ma2020acoustic}, etc., although they are less parallelizable than Transformer \cite{vaswani2017attention}.

Due to the high demand of AI model deployment on edge devices with limited power and memory, designing efficient models with low computation cost while maintaining high performance is desirable. A variety of efforts have been made in this direction. Dey and Salem \cite{dey2017gate} propose efficient GRU variants, which reduce the size of the gate matrix by adjusting the calculation method of the gate, e.g, calculating the gate vector with only hidden states as input. The Light-Gated-GRU (Li-GRU) is proposed by  \cite{ravanelli2018light}, which removes the reset gate and developes a single-gate RNN. Batch normalization is also used to optimize the model performance. Amoh and Odame \cite{amoh2019optimized} propose the Embedded Gated Recurrent Unit, which has only one gate with the Single Gate Mechanism. Simliar to Li-GRU, Fanta et al. \cite{fanta2020sitgru} discard the reset gate of GRU and replace the activation function Tanh with Sigmoid in their proposed SITGRU. Zhang et al. \cite{zhang2018simplifying} compress RNN models with a twin-gated mechanism. Although these methods can effectively reduce the number of parameters of the model, reducing the number of gate matrices may undermine the exploration of contextual information.

\begin{figure}[t]
	\centering
	\includegraphics[width=0.65\linewidth]{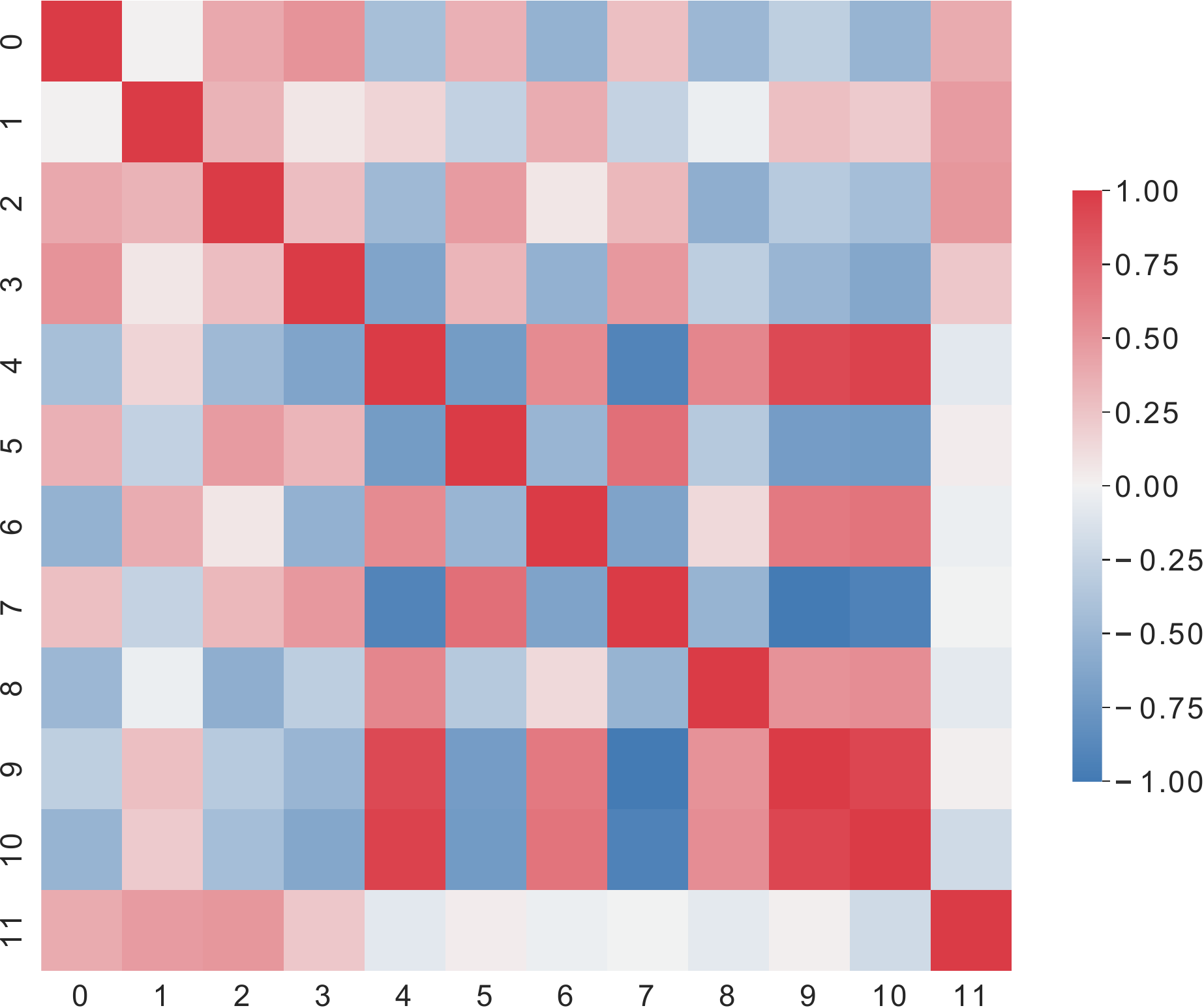}
	\caption{The cosine similarity matrix of RNN hidden states (first 12 dimensions) in a well-trained DCCRN\_GRU with 128 hidden units.}
	\label{fig:f1}
\end{figure}

In this work, we have empirically observed redundancy in hidden states of RNN models in addition to that in gate matrices studied in aforementioned previous works. We hence propose to fully explore 
 the redundancy of hidden states to compress RNNs, which has not been investigated for speech tasks. Particularly, we consider SE task with RNN models and analyze the hidden states of RNN layers. The feature map of hidden states with $m$ hidden units after $n$ time steps can be represented as $states \in \mathbb{R}^{m \times n}$, and the feature map for the $i_{th}$ hidden unit can be represented as $state_i \in \mathbb{R}^{1 \times n}$. Firstly, we perform a singular value decomposition on the feature map of hidden states from a well trained DCCRN\_GRU model. We find that only half of the singular values can reach 99\% contribution rate of principal component analysis (PCA), which indicates some  $state_i$ are relatively redundant. In addition, as shown in Figure~\ref{fig:f1} where the cosine similarities of $state_i$ at different indexes are calculated , some similarity values are relatively large, indicating a high similarity \cite{han2020ghostnet}.

Inspired by the above observation, we propose GhostRNN to reduce the redundancy in hidden states and thus construct efficient RNN models. In particular, a small part of RNN hidden states are generated by the vanilla RNN model to serve as intrinsic states. Next, cheap operations including simple linear transformation and activation function, are implemented to generate ghost states based on the intrinsic states. The ghost states are then concatenated with the intrinsic states to serve as the complete feature representation of previous time step and be passed to the next time step for further calculation. Compared to the vanilla RNN matrix multiplication, our cheap operations have fewer FLOPs and parameters, so the computation and memory costs of the GhostRNN model are significantly reduced. 
 
Experiments on the KWS and SE tasks are conducted to examine the effectiveness of our proposed method. Experimental results demonstrate that our method achieves a 0.1\% accuracy improvement on the Google Speech Commands dataset while compressing the parameters of baseline model by 40\%. In the SE task, our method improves SDR and Si-SDR by approximately 0.1dB with around 40\% compression rate.

\section{Proposed Method}
In this section, we elaborate the proposed GhostRNN with details in model compression. Without loss of generality, we use GRU to illustrate the definition of GhostRNN.  Our method can be applicable to other RNNs, e.g., LSTM. 

\subsection{RNN}
RNNs are a class of model structures that utilize hidden states to store and leverage contextual information \cite{bengio1994learning}, including popular variants GRUs and LSTMs. As one of the most commonly used RNN models, GRU is a simplified version of the LSTM, which is defined as follows:
\begin{equation}
	\bm{r}_{t} = \sigma\left(\bm{W}_{i r} \bm{x}_{t}+\bm{b}_{i r}+\bm{W}_{h r} \bm{h}_{(t-1)}+\bm{b}_{h r}\right) 
	\label{equation:eq1}
\end{equation}
\begin{equation}
	\bm{z}_{t}=\sigma\left(\bm{W}_{i z} \bm{x}_{t}+\bm{b}_{i z}+\bm{W}_{h z} \bm{h}_{(t-1)}+\bm{b}_{h z}\right) 
	\label{equation:eq2}
\end{equation}
\begin{equation}
	\bm{c}_{t}=\tanh \left(\bm{W}_{i c} \bm{x} t+\bm{b}_{i c}+\bm{r}_{t} *\left(\bm{W}_{h c} \bm{h}_{(t-1)}+\bm{b}_{h c}\right)\right)
	\label{equation:eq3}
\end{equation}
\begin{equation}
	\bm{h}_{t}=\left(1-\bm{z}_{t}\right) * \bm{c}_{t}+\bm{z}_{t} * \bm{h}_{(t-1)} 
	\label{equation:eq4}
\end{equation}
where \(\bm{r}_{t}\), \(\bm{z}_{t}\) and \(\bm{c}_{t}\) are the reset gate, update gate and candidate vector respectively. \(\bm{h}_{t}\) and \(\bm{h}_{t-1}\) are hidden states at time step \({t}\) and time step \({t-1}\), and \(\bm{x}_{t}\) is the input feature at time step \({t}\). As shown in Eq. (\ref{equation:eq1}) to (\ref{equation:eq4}), six matrices are involved in computation, among which \(\bm{W}_{i r}\), \(\bm{W}_{i z}\) and \(\bm{W}_{i c}\) have the same size, \(\bm{W}_{i r}\), \(\bm{W}_{i z}\) and \(\bm{W}_{i c}\) are also with the same size. In addition, it's worth noting that the size of all the matrices of a GRU model is closely related to both the dimension of its hidden states and input features. Consequently, to compress the number of parameters of GRU, reducing the dimensionality of hidden states is an effective way. Also as mentioned in Section~\ref{intro}, it is feasible to decrease the number of gating matrices by reducing the number of required gating vectors, which ultimately leads to the reduction in the number of parameters of the model.

\subsection{GhostRNN}

Aiming at reducing hidden state redundancy, our proposed GhostRNN employs extremely low-cost transformations to generate ghost states from intrinsic states.

\vspace{0.5em}
\noindent\textbf{Observation on the redundancy of hidden states}
\vspace{0.2em}

\begin{figure}[t]
	\centering
	\begin{subfigure}{0.72\linewidth}
		\includegraphics[width=\textwidth]{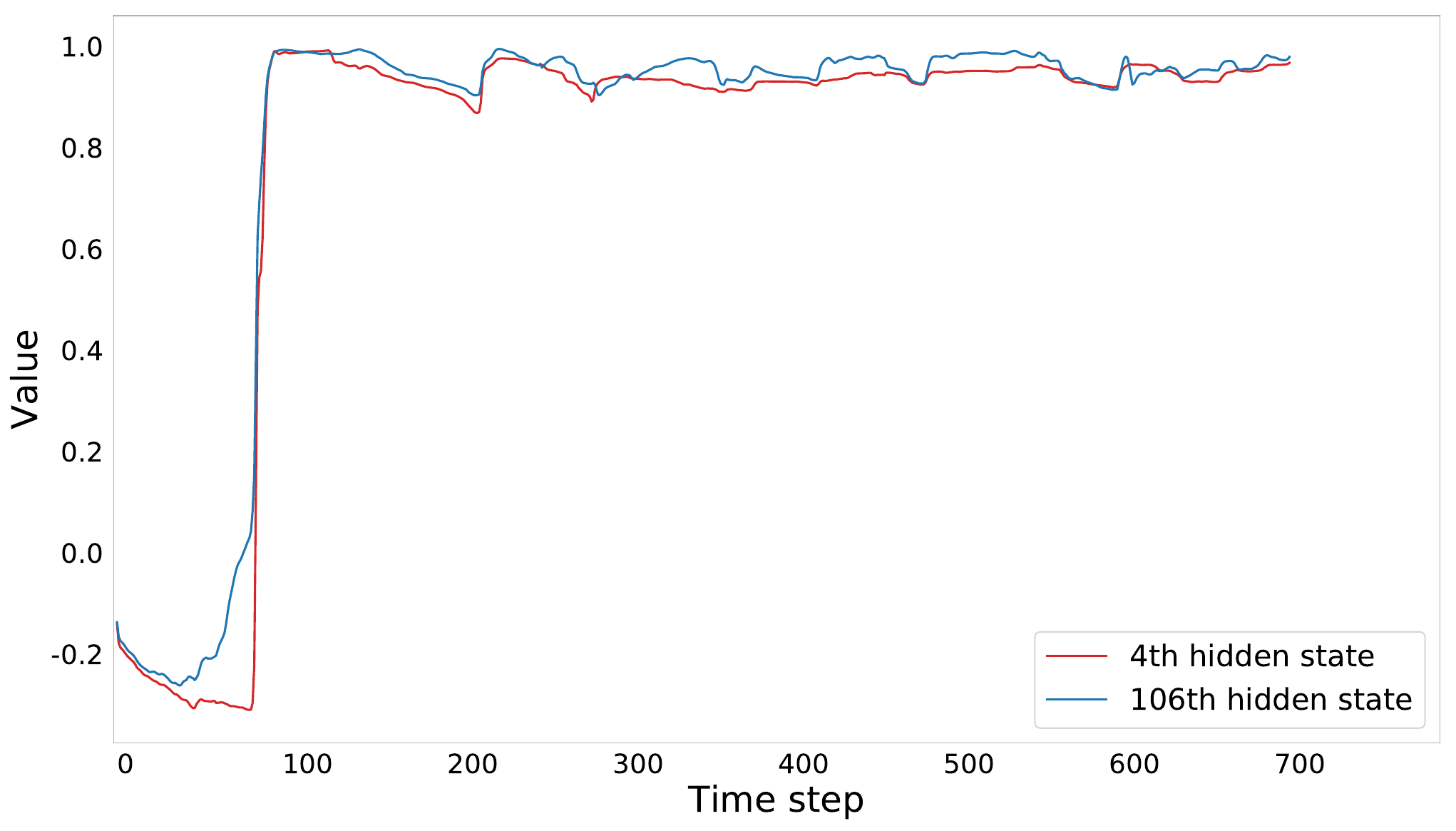}
		\caption{Value of the 4th hidden state and 106th hidden state}
		\label{fig: same}
	\end{subfigure}
	%	\hfill 
	\begin{subfigure}{0.72\linewidth}
		\includegraphics[width=\textwidth]{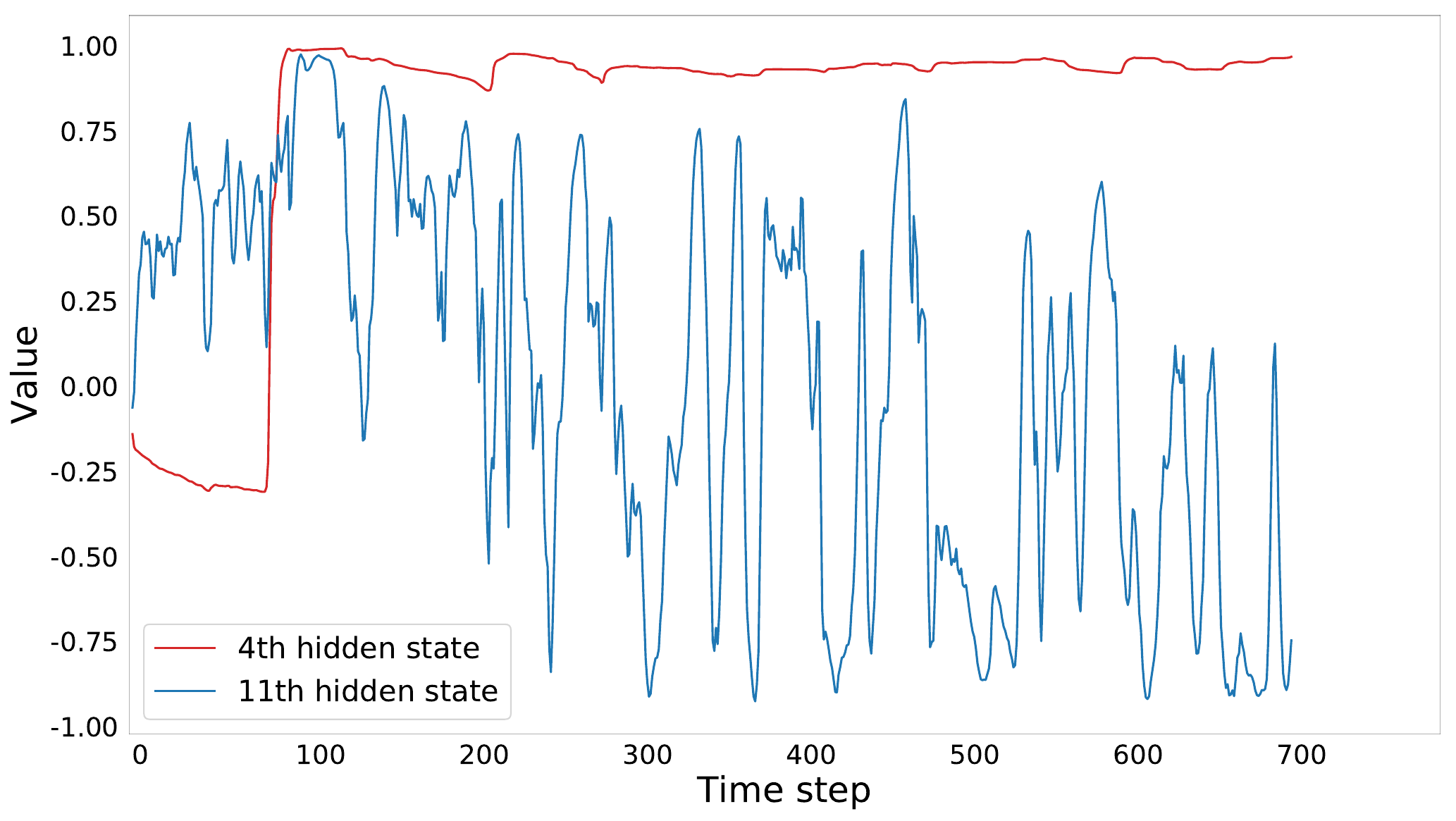}
		\caption{Value of the 4th hidden state and 11th hidden state}
		\label{fig:different}
	\end{subfigure}
	\caption{The value of hidden states at different indexes.}
	\label{fig: weightdist}
\end{figure}

As discussed in Section~\ref{intro}, previous studies usually focus on decreasing  the number of gate matrices for model compression while the redundancy of hidden states is seldom investigated, which is also crucial to effectively reduce the number of model parameters. Therefore, the redundancy of hidden states is thoroughly under investigation in this section. Initially, the PCA contribution rate is adopted as the evaluation metric. The accumulation of hidden state vectors over time is considered as a feature map and the singular value decomposition is performed on it. Based on the result, only approximately half of the singular values are necessary and the feature map can reach a 99\% PCA contribution rate, which indicates that with only about half of hidden states, the complete feature information is possible to be constructed. Furthermore, as shown in Figure \ref{fig:f1}, we take the accumulation of hidden states of different dimensions over time steps as the state components $state_i$ and calculate the cosine similarity between different components. It shows that the cosine similarity of certain components  nearly approaches 1.0. A group of components with highly correlated trends is shown in Figure \ref{fig: same}, which indicates that hidden states of GRU contain redundancy. On the other hand, as shown in Figure \ref{fig:different}, the cosine similarity between specific components is close to 0, indicating that these state components are almost orthogonal and hence necessary and irreplaceable. Therefore keeping the necessary state components and eliminating redundant components is a straightforward and practical way to compress the RNN models.

\vspace{0.5em}
\noindent\textbf{GhostRNN module}
\vspace{0.2em}

\begin{figure}[t]
	\centering
	\begin{subfigure}[b]{0.75\linewidth}
		\includegraphics[width=\textwidth]{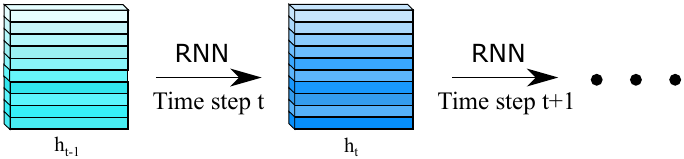}
		\caption{RNN}
		\label{fig: GRU}
	\end{subfigure}
	\hfill 
	\begin{subfigure}[b]{1.0\linewidth}
		\includegraphics[width=\textwidth]{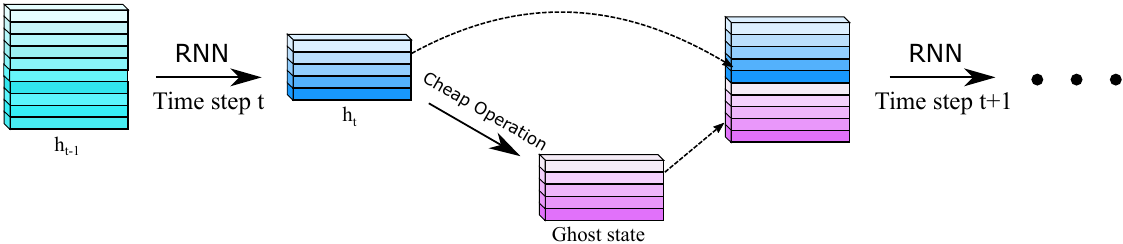}
		\caption{GhostRNN}
		\label{fig:GhostRNN}
	\end{subfigure}
	\caption{Schematic diagram of the RNN layer and proposed Ghost RNN.}
	\label{fig: Ghoststate}
\end{figure}

Based on the above results and analysis, we propose the GhostRNN module to construct the ghost states based on the intrinsic states as shown in Figure \ref{fig: Ghoststate}, which can be defined as follows:

\begin{equation}
	\bm{r}_{t}=\sigma\left(\bm{W}_{i r} \bm{x}_{t}+\bm{b}_{i r}+\bm{W}_{h r} [\bm{h}_{(t-1)} \ \bm{g}_{(t-1)}]+\bm{b}_{h r}\right) 
	\label{equation:eq5}
\end{equation}
\begin{equation}
	\bm{z}_{t}=\sigma\left(\bm{W}_{i z} \bm{x}_{t}+\bm{b}_{i z}+\bm{W}_{h z} [\bm{h}_{(t-1)} \ \bm{g}_{(t-1)}]+\bm{b}_{h z}\right) 
	\label{equation:eq6}
\end{equation}
\begin{align}
	\notag
	\bm{c}_{t} =& \tanh \left(\bm{W}_{i c} \bm{x}_t+\bm{b}_{i c}+\bm{r}_{t} *\left(\bm{W}_{h c} \bm{h}_{(t-1)}+\bm{b}_{h c}\right) \right.\\
	 &\left. + \bm{W}_{gc}\bm{g}_{t-1} + \bm{b}_{gc} \right)
	\label{equation:eq7}
\end{align}
\begin{equation}
	\bm{h}_{t}=\left(1-\bm{z}_{t}\right) * \bm{c}_{t}+\bm{z_{t}} * \bm{h}_{(t-1)} 
	\label{equation:eq8}
\end{equation}
\begin{equation}
	\bm{g}_{t}=\phi\left(\bm{h}_{t}\right)
	\label{equation:eq9}
\end{equation}
where \(\bm{g}_{t}\) and \(\bm{g}_{t-1}\) are the ghost states of the GRU model at time step \({t}\) and time step \({t-1}\) generated by the original intrinsic states \(\bm{h}_{t}\) and \(\bm{h}_{t-1}\) through simple linear transformation operations and activations denoted by \(\phi\).  The intrinsic states at time step  \({t}\)
are obtained by following Eqs (\ref{equation:eq5}) to (\ref{equation:eq8}).
 \([ \ ]\) represents the concatenation operation to concatenate intrinsic and ghost states. The concatenated states are used as the input for the next time step. Specifically, at time step \({t}\) the GhostRNN receives both the previous hidden state \(\bm{h}_{t-1}\) and corresponding ghost state \(\bm{g}_{t-1}\) synchronously with the current input feature \(\bm{x}_{t}\).  This process is repeated to complete the calculation of the GhostRNN. As mentioned above, the GhostRNN method, which leverages the redundancy of  hidden states, can be applied to compress all RNN models.

\subsection{Analysis on the Number of Parameters and Computational Complexity}
The number of parameters of vanila GRU and our proposed GhostRNN can be calculated as follows:

\begin{equation}
	\rm{Param_{GRU}}=3\left(\rm{Dim_{feature}} + \rm{Dim_{state}}\right) \rm{Dim_{state}} 
	\label{equation:eq10}
\end{equation}
%
%\begin{equation}
\begin{align}
	\notag
	\rm{Param_{GhostRNN}}=&3\left(\rm{Dim_{feature}} + \rm{Dim_{state}}\right) 
	\left(\rm{Dim_{state}} / r \right) \\
	 &+ \rm{Param_{\phi}}   
	\label{equation:eq11}
\end{align}
%\end{equation}
\begin{align}
	\notag 
	\rm{Param_{\phi}} &= \left(\rm{Dim_{state}} / r \right) \left(\rm{Dim_{state}} - \left(\rm{Dim_{state}} / r \right) \right) \\
	&= \rm{Dim_{state}}^{2} \left( (r-1) / r^{2} \right)
	\label{equation:eq12}
\end{align}
where \(r\) represents the ratio of the complete state to the intrinsic state. As shown in Eqs (\ref{equation:eq10}) to (\ref{equation:eq11}), the dimension of output hidden states in GhostRNN is divided by the ratio \(r\) compared with those in the vanila GRU. Although an additional cheap operation module consisting of linear layers is applied, according to Eq. (\ref{equation:eq12}) the number of parameters of the cheap operation module is far smaller than that of the GRU model. As a result, the total number of parameters of the GhostRNN will be compressed by the factor \(r\). If the cheap operation module is constructed by other calculation methods such as vanilla linear transformation without parameters, the compression ratio can reach up to \(r\). As for the computational complexity, since all the matrices in GRU are linear layers, the computational complexity of GRU is almost proportional to the number of parameters. Thus, the computational complexity of GhostRNN will also be reduced by the same factor \(r\).

\begin{table}[h]
	\vspace{-10pt}
	\caption{Overall performance of KWS}
	\label{tab:system performance}
	\centering
	\vspace{-10pt}
	\begin{tabular}{ l c c c}
		\toprule
		\multicolumn{1}{c}{\textbf{System}} &
		\multicolumn{1}{c}{\textbf{\# Params}} &
		\multicolumn{1}{c}{\textbf{\# MACs}} &
		\multicolumn{1}{c}{\textbf{Accuracy (\%)}}\\
		\midrule
		GRU & 498K & 24.1M & 94.68 \\
		GRU & 295K & 14.2M & 94.49 \\
		Li-GRU & 334K& 16.1M & 93.49 \\
		SITGRU & 334K & 16.1M & 94.26 \\
		GhostRNN & 292K & 14.0M & \textbf{94.79} \\
		\bottomrule		
	\end{tabular}
	\vspace{-10pt}
\end{table}

\section{Experiments}

Experiments on two tasks were conducted to evaluate the effectiveness of our method: KWS and SE.

\subsection{Datasets}

The Google Speech Commands dataset v0.02 \cite{warden2018speech} which contains thousands of one-second audio samples divided into 30 categories was used in our experiments for KWS. Following the previous work \cite{zhang2017hello, mekonnen2022end}, 12 categories were selected: "yes," "no," "up," "down," "left," "right," "on," "off," "stop," "go," silence, and unknown. We used 36,923, 4,445, and 4,890 of these samples for training, validation, and testing, respectively. The 10-dimensional Mel-frequency cepstral coefficients (MFCC) were used as the speech feature with a window length of 40 ms and a window shift of 20 ms which results in a feature map of size 49x10 for each speech sample and the data augmentation techniques such as adding background noise and random shift as suggested in \cite{zhang2017hello} were employed to enhance the roubustness of the models.

To evaluate the performance of our method on the SE task, we used the LibriMix dataset \cite{cosentino2020librimix} which generates noisy speech clips by combining clean speech from LibriSpeech \cite{panayotov2015librispeech} and noise from the WHAM! dataset \cite{wichern2019wham}. We used the 16 kHz version of the train-360 data with a total of 50,800 training samples, 3000 validation samples and 3,000 testing samples, resulting in a total of 234 hours of data. We implemented the same data preprocessing as that in Asteroid \cite{Pariente2020Asteroid}.

\subsection{Settings and Evaluation Metrics}

\begin{table*}[!h]
	\caption{Performance of DCCRN, DCCRN\_Ghost, GRU-TasNet and GhostRNN-TasNet }
	\label{tab:gru-tasnet performance}
	\centering
	\begin{threeparttable}
	\begin{tabular}{l c c c c c c c}
		\toprule
		\multicolumn{1}{c}{\textbf{System}} &
		\multicolumn{1}{c}{\textbf{\# Parameters}} &
		\multicolumn{1}{c}{\textbf{\# RNN MACs}\tnote{1} } &
		\multicolumn{1}{c}{\textbf{SDR (dB)}}&
		\multicolumn{1}{c}{\textbf{SDRi (dB)}}&
		\multicolumn{1}{c}{\textbf{Si-SDR (dB)}}&
		\multicolumn{1}{c}{\textbf{Si-SDRi (dB)}}&
		\multicolumn{1}{c}{\textbf{STOI (\%)}}\\
		\midrule
		DCCRN\_GRU128 & 3.4M & 0.7M & 13.99 & 10.49 & 13.47 & 10.02 & 91.6  \\
		DCCRN\_GRU80 & 3.1M & 0.4M & 13.89 & 10.39 & 13.36 & 9.91 & 91.4  \\
		DCCRN\_Ghost128 & 3.1M & 0.4M  &  \textbf{13.93} &  \textbf{10.43} &  \textbf{13.40} &  \textbf{9.95} &  \textbf{91.5}  \\
		\hline
		GRU512-TasNet & 5.6M & 4.7M & 13.14 & 9.64 & 12.63 & 9.18 & 89.5  \\
		GRU384-TasNet & 3.4M & 2.8M & 13.09 & 9.59 & 12.56 & 9.11 & 89.4  \\
		GhostRNN512-TasNet & 3.4M & 2.6M & \textbf{13.26} & \textbf{9.76} & \textbf{12.73} & \textbf{9.28} & \textbf{89.7}  \\
		GRU192-TasNet & 1.6M & 0.8M & 12.82 & 9.31 & 12.28 & 8.83 & 88.9  \\
		GRU136-TasNet & 1.2M & 0.5M & 12.48 & 8.98 & 11.92 & 8.47 & 88.1  \\
		GhostRNN192-TasNet & 1.2M & 0.5M & \textbf{12.61} & \textbf{9.10} & \textbf{12.05} & \textbf{8.60} & \textbf{88.4}  \\
		\bottomrule
	\end{tabular}
	\begin{tablenotes}
		\footnotesize
		\item[1] The MACs of RNN with single time step is presented to show the difference in computation cost clearly. 
		\end{tablenotes}
	\end{threeparttable}
		
\end{table*}

During training of KWS, the standard cross-entropy loss and the Adam optimizer with a batch size of 100 were employed. We used a step-down learning rate strategy, where the initial learning rate was 5e-4 with the step setting [10,000,20,000]. All the models were trained from scratch for a total of 30,000 iterations and evaluated by the accuracy metric \cite{zhang2017hello}. To ensure the reliability of our results, each model was trained with the same configuration for three times and the average experiment results are reported here.

During the training process of SE, the permutation invariant loss and the ADAM optimizer were used and the batch size was set to 12 for DCRNN \cite{hu2020dccrn} and 32 for GRU-TasNet \cite{luo2018real}. The learning rate decay strategy and early stopping strategy were both applied in all experiments. The initial learning rate was set to 0.001 and a weight decay of 1e-5 was applied. In terms of the filterbank, our settings were consistent with those described in \cite{Pariente2020Asteroid}. For evaluation, 5 metrics, namely Signal-to-Distortion Ratio (SDR), SDR improvements (SDRi), Scale-Invariant Signal-to-Distortion Ratio (Si-SDR) \cite{le2019sdr}, Si-SDR improvements (Si-SDRi), and Short-Time Objective Intelligibility (STOI) \cite{taal2011algorithm} were used.

\subsection{Baselines}
\noindent\textbf{KWS model}
\vspace{0.2em}

As demonstrated in Table \ref{tab:system performance}, a GRU model of approximately 500k in size and another GRU model of 295k in size were employed as baseline models for KWS \cite{zhang2017hello}, while Li-GRU \cite{ravanelli2018light} and SITGRU \cite{fanta2020sitgru} were implemented for comparison.

\vspace{0.5em}
\noindent\textbf{SE models}
\vspace{0.2em}

Two types of models were selected as the baselines model for comparison for the SE task. The first one is the DCCRN \cite{hu2020dccrn}, which is mainly based on convolution layers and supplemented with RNN layers. The second one is the GRU-TasNet \cite{luo2018real}, in which RNN is the primary module. A brief introduction of them is provided below:

\begin{itemize}
	\item DCCRN. This model consists of three main components: a convolution encoder, a transpose convolution decoder and an RNN module. In our experiment, we chose the DCCRN-CL \cite{hu2020dccrn} model and replaced the LSTM with GRU, in which two sizes of the hidden unit 128 and 80 were chosen to construct the different baseline models with different size.
	
	\item GRU-TasNet. This model is optimized by the Time-domain Audio Separation Network, which consists of three parts: a 1-D convolutional encoder, a 1-D deconvolutional decoder, and a Deep LSTM separation module \cite{luo2018real}. In our experiments, the LSTM  was replaced with GRU. Four baseline models with different sizes were designed, differing in the hidden size of the GRU: 512, 384, 192, and 136.
\end{itemize}

\subsection{Results on KWS}

Table \ref{tab:system performance} presents the experiment results and model parameters. Our proposed GhostRNN model is compared with two baselines of 500k GRU and 300k GRU, as well as two other model compression methods: Li-GRU \cite{ravanelli2018light} and SITGRU \cite{fanta2020sitgru}. The results show that our GhostRNN with about 40\% fewer parameters achieves approximately a 0.1\% improvement on the accuracy rate over the 500k GRU model and also outperforms the Li-GRU and SIT-GRU models with slightly more parameters, which indicates the effectiveness of our proposed GhostRNN.

\subsection{Results on SE}

Table \ref{tab:gru-tasnet performance} presents the results of DCCRN and DCCRN\_Ghost128 on the librimix1 dataset. The results indicate that pruning the hidden size of the GRU layer in the DCCRN model by 10\% for model compression leads to a decrease of approximately 0.1 dB in both SDR and Si-SDR metrics. In contrast, when applying our proposed GhostRNN compression method and compressing approximately 10\% of the parameter, the SDR and Si-SDR metrics only decrease by approximately 0.05 dB. These findings clearly demonstrate the effectiveness of GhostRNN.

Table \ref{tab:gru-tasnet performance} presents the results of GRU-TasNet and GhostRNN-TasNet on the librimix1 dataset. The metrics show a slight decrease when the model is compressed from GRU512-TasNet to GRU384-TasNet, indicating that GRU512-TasNet has redundant parameters. In this case, the GhostRNN method yields a performance improvement of over 0.1 dB in SDR and Si-SDR, with 40\% fewer parameters than GRU512-TasNet. However, when the model is further compressed from 1.6M (GRU192-TasNet) to 1.2M (GRU136-TasNet), the performance drops noticeably, demonstrating that the model has low redundancy. In this scenario, GhostRNN192 has an advantage of approximately 0.13 dB in SDR and Si-SDR compared to GRU136-TasNet. In summary, GhostRNN is an effective compression method for RNN models.

\section{Conclusions}

In this paper, we have presented GhostRNN for RNN model compression based on the observation of the redundancy in hidden states. In our GhostRNN, given the intrinsic hidden states, the extreme low-cost transformation layers are applied to generate the ghost states which significantly reduces the number of parameters and the computation cost of the vanllia GRU model but achieves competitive performance. Experimental results demonstrate that our method achieves a 0.1\% accuracy improvement on the Google Speech Commands dataset while compressing the parameters of baseline model by 40\%. In the SE task, our method improves SDR and Si-SDR by approximately 0.1 dB with around 40\% compression rate. Additionally, our method outperforms the GRU based model with the same number of parameters by approximately 0.13 dB in terms of SDR and other evaluation metrics. Overall, the proposed GhostRNN is a simple yet effective method for RNN model compressing.
In the future work, it is worth to investigate the extension of GhostRNN to other RNN structures, such as LSTM, and further explore novel ghost state generation methods to achieve better balance on the reduction of the model computational complexity and performance. Additionally, we plan to explore the potential benefits of combining GhostRNN with other existing RNN compression techniques.

\section{Acknowledgements}
Deyi Xiong was partially supported by the Natural Science Foundation of Xinjiang Uygur Autonomous Region (No. 2022D01D43). We gratefully acknowledge the support of MindSpore \cite{mindspore}, CANN(Compute Architecture for Neural Networks) and Ascend AI Processor used for this research. We would like to thank the anonymous reviewers for their insightful comments.

\bibliographystyle{IEEEtran}
\bibliography{mybib}

\end{document}